\documentclass{article}

\usepackage[dblblindworkshop, final]{neurips_2026}
\workshoptitle{Personalized, Aligned, Long-Term Memory for AI Systems}

\usepackage[utf8]{inputenc}
\usepackage[T1]{fontenc}
\usepackage{hyperref}
\usepackage{url}
\usepackage{booktabs}
\usepackage{amsmath}
\usepackage{graphicx}
\usepackage{microtype}
\usepackage{xcolor}
\usepackage{enumitem}
\usepackage{multirow}
\usepackage{longtable}
\usepackage{array}
\usepackage{adjustbox}
\usepackage{caption}
\usepackage{float}
\usepackage{placeins}
\usepackage{pgfplots}
\pgfplotsset{compat=1.18}

\title{When Retrieval Is Not Enough: Obsolete Facts in Agent Memory}
\author{%
  Oleeviya Babu Poikarayil$^{1}$\quad
  Tony Basil Babu$^{2}$\quad
  Abdulrahman Nahhas$^{3}$\quad
  Het Darshan Mehta$^{3}$ \\
  \textbf{Jawid Ahmad Baktash}$^{4}$\quad
  \textbf{Ernesto William De Luca}$^{3}$\quad
  \textbf{Klaus Turowski}$^{3}$ \\[4pt]
  $^{1}$Faculty of Computer Science, Otto-von-Guericke-Universit\"at Magdeburg, Germany \\
  $^{2}$University of Kerala, India \\
  $^{3}$Otto-von-Guericke-Universit\"at Magdeburg, Germany \\
  $^{4}$Technical University of Munich, Germany \\[4pt]
  \texttt{oleeviyazehr@gmail.com, tonybasil1233@gmail.com} \\
  \texttt{abdulrahman.nahhas@ovgu.de, het.mehta@ovgu.de} \\
  \texttt{jawid.baktash@tum.de, ernesto.deluca@ovgu.de, klaus.turowski@ovgu.de}
}

\begin{document}
\maketitle
\begin{abstract}
Memory-agent benchmarks often collapse retrieval and answer generation into one score, making it hard to tell whether a failure comes from missing an update or from not using an update that was already retrieved. We examine this distinction on the single-hop FactConsolidation task of MemoryAgentBench. With Qwen2.5-7B-Instruct on 6K-history BM25 accuracy drops from 45\% to 32\%  when the retrieved evidence budget grows from 64 to 512 tokens, even though a gold-assisted post-hoc audit finds the newest statement in every retrieved set. Most of the errors occur on questions that also contain an obsolete statement version. Under a frozen prompt and equal four-statement evidence size, replacing one unrelated statement with the obsolete competing statement reduces accuracy from 100\% to 44.6\% for Qwen2.5 and from 100\% to 87.8\% for Qwen3-14B, with all 50 harmful paired flips classified as stale. A 32K-history replication likewise favors BM25-64 over BM25-512 (69\% versus 58\%). These results suggest that retrieving the correct fact is not enough. An obsolete competing statement can be more harmful than an unrelated piece of evidence of the same size.
\end{abstract}

\section{Introduction}
Persistent memory lets an agent carry preferences, task state, and factual updates across sessions. A deployed system must therefore solve at least two distinct problems: retrieving the relevant information, and selecting the currently valid version when memory contains both an old fact and a later correction. End-to-end accuracy hides this distinction. The retriever may return the newest fact, yet the language model can still answer with the older one.

We ask the following question: \emph{when the correct update is retrieved, how do evidence budget, unrelated distractors, and an older competing fact affect whether an LLM uses it?} We study this question using the MemoryAgentBench's FactConsolidation task \citep{hu2026memoryagentbench}. In this task, numbered statements are chronological and higher numbers replace lower conflicting statements. The facts are synthetic and can therefore contradict ordinary world knowledge. For example, the memory may first state that the author of \emph{The Marriage of Figaro} is Pierre Beaumarchais and later state that it is Thomas Kyd. Under the benchmark's chronology, Thomas Kyd is then the correct answer.

We address the question in three parts: \textbf{RQ1} asks how the evidence budget affects the use of a retrieved update; \textbf{RQ2} asks whether failures are concentrated on questions for which an obsolete value is also retrieved; and \textbf{RQ3} asks whether an obsolete competing fact is more harmful than unrelated evidence of the same size.

We vary the BM25 budget, compare questions with and without an obsolete competitor, and run an equal-size substitution on two models, with a 32K budget replication. Together, these test a focused claim: \textbf{obsolete target evidence is substantially more damaging than unrelated evidence of the same size.} Appendix~\ref{app:inventory} provides a complete inventory of the exploratory, diagnostic, final, replication, and excluded runs.

\paragraph{Relation to prior work.}
Memory benchmarks such as LoCoMo, LongMemEval, and MemoryAgentBench cover multi-session interaction and temporal updates \citep{maharana2024locomo,wu2025longmemeval,hu2026memoryagentbench}, while RAG/memory systems provide compact evidence \citep{lewis2020rag,chhikara2025mem0}. Long-context studies document distraction and contradictory-evidence effects \citep{shi2023distracted,liu2024lost,kurfali2025conflicting}; recent work also targets temporal or invalidated knowledge \citep{wallat2026facts,chao2026stale}. Most closely, \citet{reddy2026freshness} study FactConsolidation but note that their pipeline comparison jointly changes resolver, prompt, format, and temperature. We extend their pipeline-level finding with a controlled intervention. It isolates evidence composition while holding reader, prompt, latest fact, and four-statement size fixed.

\section{Experimental design}
\paragraph{Data and chains.}
We use the \texttt{Conflict\_Resolution} split of MemoryAgentBench, focusing on \texttt{factconsolidation\_sh\_6k} and \texttt{factconsolidation\_sh\_32k}. The 6K history contains 455 statements and 100 questions. Template-based chain reconstruction resolves all 100 questions: 74 have one older conflicting statement and 26 have no previous conflict. The 32K history contains 2,310 statements and 100 questions. We use the benchmark's normalized substring metric for the primary results and manually inspect semantic scoring edge cases.

\paragraph{Retrieval budgets.}
We rank individual statements with BM25 \citep{robertson2009bm25}. We add them until reaching a token budget, and restore chronological order. The 6K sweep uses budget of 64, 128, 256, and 512 evidence tokens; the 32K replication uses the two endpoints. A post-hoc support audit uses the benchmark answer only to check whether the intended latest statement is present in the retrieved evidence. It is therefore a diagnostic, not a deployable retrieval procedure.

\paragraph{Equal-size intervention.}
For each of the 74 conflict questions, we build two four-statement contexts: (A) the latest target statement and three unrelated statements, and (B) the obsolete target statement, the latest statement, and two of the same unrelated statements. The distractors are selected deterministically, have zero content-word overlap with the question, and exclude update-chain values. Thus A and B differ only by replacing one unrelated statement with the obsolete competitor. These 74 questions were also used for prompt development and the pilot. This is a within-cohort paired comparison rather than held-out confirmation. Freezing the prompt across conditions protects the A/B contrast.

\paragraph{Models and inference.}
We evaluate \texttt{qwen2.5:7b-instruct} and \texttt{qwen3:14b} \citep{qwen2024qwen25,qwen2025qwen3}, served through Ollama on a university GPU system. Local Python scripts send requests to the HTTP chat API over a VPN. The final controlled experiments and corrected 32K replication use temperature 0, seed 42, \texttt{num\_ctx}=2048, and \texttt{num\_predict}=64; the Qwen3 run additionally sets \texttt{think=false}. Earlier 6K runs use larger context settings (Appendix~\ref{app:repro}). The frozen final prompt instructs to use only supplied numbered memories, treat higher-numbered conflicting statements as newer replacements, and return the shortest answer phrase. We use paired exact McNemar tests as benchmark diagnostics; questions share a history, so we avoid population-level interpretations.

\section{Results}
\subsection{Budget growth exposes a conflict-resolution failure}
On the 6K history, BM25 accuracy decreases monotonically from 45\% at 64 tokens to 32\% at 512 tokens (Table~\ref{tab:budget}). A gold-assisted post-hoc audit identifies the latest statement in all 100 retrieved sets at every budget. The decline in accuracy therefore cannot be explained simply by loss of the intended update. In the paired 64-versus-512 comparison, 14 questions are correct only at 64 tokens, while just one is correct only at 512 ($p=9.77\times10^{-4}$).

\begin{table}[t]
\centering
\caption{Evidence-budget results. ``Conflict'' uses reconstructed chains. Gold-assisted support is chain-based at 6K and direct literal benchmark-gold presence at 32K after two ambiguous chain labels. The validated 32K conflict subset has 63/63 chain support at both budgets.}
\label{tab:budget}
\small
\begin{tabular}{llrrrr}
\toprule
History & Budget & Accuracy & Conflict acc. & Gold/latest support & Mean stmts.\\
\midrule
\multirow{4}{*}{6K} & 64  & \textbf{45} & \textbf{25.7} & 100/100 & 4.17\\
& 128 & 40 & 20.3 & 100/100 & 8.57\\
& 256 & 36 & 16.2 & 100/100 & 16.99\\
& 512 & 32 & 12.2 & 100/100 & 33.60\\
\midrule
\multirow{2}{*}{32K} & 64  & \textbf{69} & \textbf{55.6} & 100/100 & 3.96\\
& 512 & 58 & 34.9 & 100/100 & 32.96\\
\bottomrule
\end{tabular}
\end{table}

Chain stratification localizes the 6K failures. No-conflict accuracy is 100\% at BM25-64 and 88.5\% at BM25-512, while conflict accuracy falls from 25.7\% to 12.2\%. Since both the obsolete and latest statements are already retrieved at every 6K budget, this decline is not caused by newly retrieving the competitor. Additional evidence appears to compound an already difficult selection. Because conflict status is a property of the question, this comparison is observational. A separate gold-assisted diagnostic reaches 97\%, but it uses a different prompt and gold-derived evidence selection. We treat it only as evidence that short, targeted contexts can be answerable. Appendix~\ref{app:baselines} reports the preliminary and gold-assisted diagnostics, Appendix~\ref{app:budget} gives the complete 6K budget analysis, and Appendix~\ref{app:conflict} gives the full conflict-stratified analysis.

\subsection{An obsolete fact is more harmful than an unrelated fact}
\begin{figure}[t]
\centering
\includegraphics[width=0.50\linewidth]
{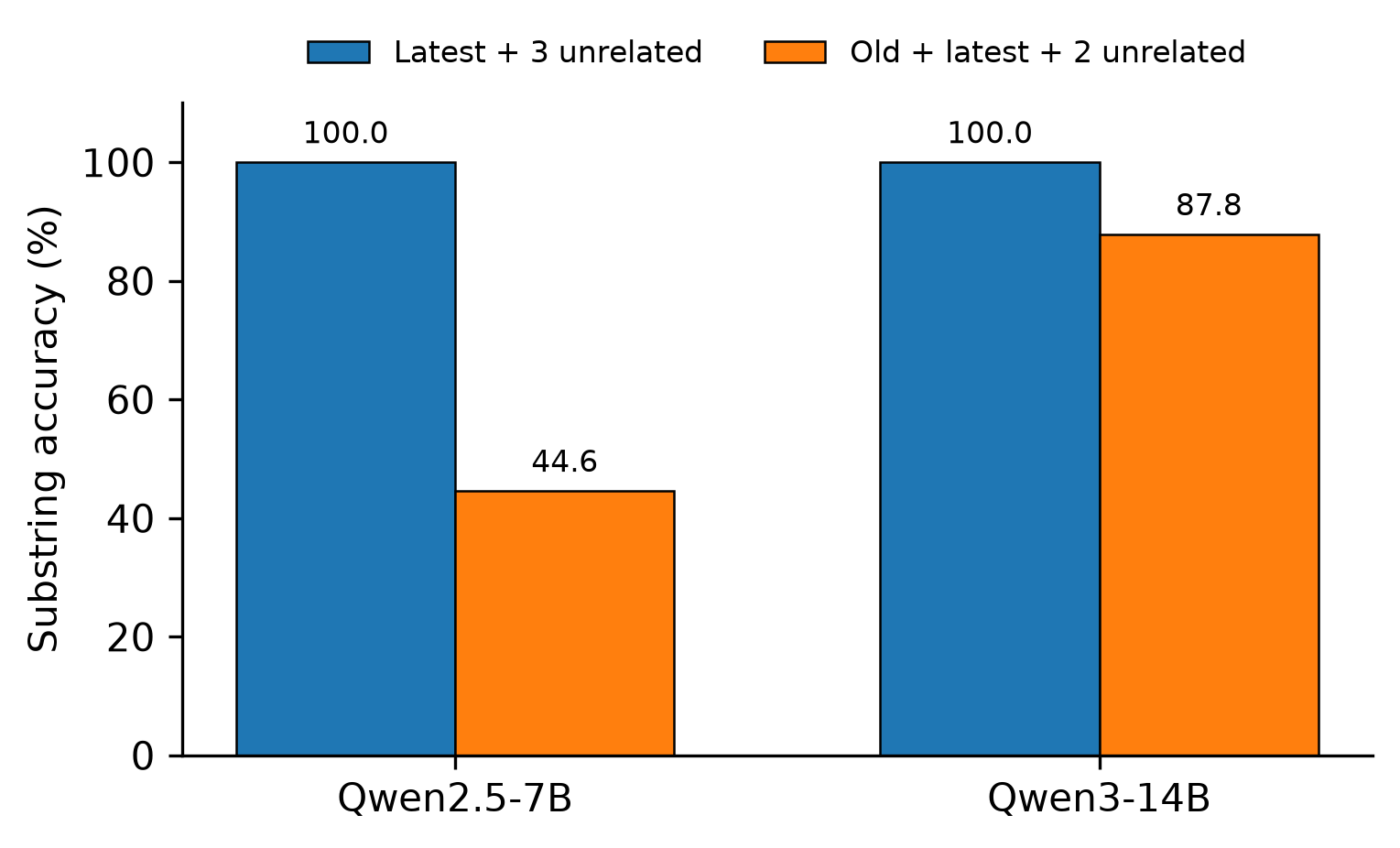}
\caption{Equal-size intervention on 74 conflict questions. Replacing one unrelated statement with the obsolete target fact harms both models.}
\label{fig:controlled}
\end{figure}

The equal-size intervention gives the central result (Figure~\ref{fig:controlled}). Qwen2.5 answers all 74 questions correctly in condition A but only 33 in condition B, a 55.4-point drop. The paired comparison has 41 questions correct only in A and none correct only in B  ($p=9.09\times10^{-13}$). All 41 harmful flips are classified as stale. Qwen3 is considerably more robust, but the same direction remains. Its accuracy falls from 100\% in condition A to 87.8\% in condition B, with nine A-only correct cases, no B-only correct cases, and nine stale selections ($p=0.0039$).

For example, both models answer ``Thomas Kyd'' for \emph{The Marriage of Figaro} in condition A, but can switch to ``Pierre Beaumarchais'' once the older statement replaces an unrelated memory, even though the newer statement remains visible and the prompt specifies the chronology. Appendix~\ref{app:exploratory} reports the earlier abstaining-prompt experiment, while Appendix~\ref{app:final} gives the frozen-prompt pilot, complete cross-model results, and all paired flips.

\subsection{32K replication and manual audit}
The larger 32K history shows the same overall direction. The official 100-question scores are 69\% for BM25-64 and 58\% for BM25-512. 18 questions are correct only at 64 tokens and seven are correct only at 512 ($p=0.0433$). All 18 harmful flips are classified as correct-to-stale. Manual inspection identified two ambiguous cases among automatically labeled no-conflict questions. We keep the official scores as primary results. Excluding only those two for sensitivity gives $n=98$, 69/98 (70.4\%) versus 57/98 (58.2\%), with 18 versus six discordant successes ($p=0.0227$).

Manual metric review found both a false negative (``goalie'' for gold ``goaltender'') and a false positive (``rugby union'' contains gold ``rugby'' despite matching the obsolete value), plus two final Qwen2.5 outputs containing both latest and stale strings. We keep the official substring scores. Appendices~\ref{app:sh32k} and~\ref{app:manual} give the full 32K replication and the manual audit.

\begin{table}[H]
\centering
\caption{Representative audited outputs. The benchmark truth is the later synthetic memory, not ordinary world knowledge.}
\label{tab:audit}
\scriptsize
\setlength{\tabcolsep}{2.5pt}
\begin{tabular}{p{0.26\linewidth}p{0.28\linewidth}p{0.18\linewidth}p{0.26\linewidth}}
\toprule
Question & Latest / obsolete & Larger or conflict answer & Audit verdict\\
\toprule
Author of \emph{The Marriage of Figaro}? & Thomas Kyd / Pierre Beaumarchais & Pierre Beaumarchais & Direct stale selection\\
\midrule
Sport associated with flanker? & rugby / rugby union & rugby union & Metric false positive; stale value\\
\midrule
Position of Lisa Leslie? & goaltender / center & goalie & Semantic match; metric false negative\\
\bottomrule
\end{tabular}
\end{table}

\section{Discussion and conclusion}

\paragraph{Answers to the research questions.}

Larger budgets lower accuracy even when the latest statement is present (\textbf{RQ1}). The loss concentrates on conflict questions (\textbf{RQ2}) and at equal size, replacing an unrelated statement with the obsolete competitor harms both models, with every harmful flip selecting the stale value (\textbf{RQ3}).

\paragraph{Evaluation and system implications.}
Retrieving an update does not guarantee that the model will use it. Evaluation should therefore distinguish whether the latest evidence is retrieved, whether obsolete evidence is also retrieved, and which version the reader selects. Evidence composition matters beyond token count. An obsolete target statement is more harmful than an unrelated statement. These results motivate memory systems that retire, down-weight, or explicitly mark superseded facts as inactive while preserving provenance and validity information during retrieval.

\paragraph{Model variation and mechanism.}
Qwen3's smaller drop suggests that model choice may affect robustness, although its nine harmful flips are still stale selections. In this benchmark, obsolete values often match familiar world knowledge while the synthetic updates contradict it. Our results therefore establish susceptibility to these competitors, not that temporal obsolescence alone drives the effect. Counterbalancing chronology or using unfamiliar entities would separate the two. Our behavioral intervention does not establish whether the cause is attention allocation, prior strength, or decoding dynamics, so we avoid a mechanistic claim.

\paragraph{Scope and broader impact.}
Our study is limited to two open-weight Qwen models, deterministic decoding, one synthetic benchmark, single-hop chains of length at most two, and lexical retrieval. We therefore do not establish the same behavior for dense retrieval, multi-hop memory, or end-to-end systems such as Mem0. Both models are from the Qwen family. Cross-family generalization is untested, and longer update chains or multi-hop dependencies may compound the effect. The gold-assisted audits are diagnostic rather than deployable retrieval evaluations, and the questions share a common benchmark history, which limits population-level statistical interpretation. Appendix~\ref{app:validity} documents the excluded runs and result-verification checks. In deployed agents, stale sensitive or operational information could lead to outdated recommendations or disclosures. Version-aware retirement can mitigate this risk. Within these boundaries, \textbf{when both versions are present, obsolete evidence can override a retrieved correction.}

\clearpage

\begin{ack}
We thank the anonymous reviewers of the PALM workshop for their constructive feedback.
Experiments were run on GPU resources provided by Otto-von-Guericke-Universit\"at Magdeburg.
\end{ack}

\bibliographystyle{plainnat}
\bibliography{references}

@inproceedings{hu2026memoryagentbench,
  author    = {Yuanzhe Hu and Yu Wang and Julian J. McAuley},
  title     = {Evaluating Memory in {LLM} Agents via Incremental Multi-Turn Interactions},
  booktitle = {The Fourteenth International Conference on Learning Representations, {ICLR} 2026},
  year      = {2026},
  url       = {https://iclr.cc/virtual/2026/poster/10010781}
}

@inproceedings{wu2025longmemeval,
  author       = {Di Wu and
                  Hongwei Wang and
                  Wenhao Yu and
                  Yuwei Zhang and
                  Kai{-}Wei Chang and
                  Dong Yu},
  title        = {LongMemEval: Benchmarking Chat Assistants on Long-Term Interactive
                  Memory},
  booktitle    = {The Thirteenth International Conference on Learning Representations,
                  {ICLR} 2025, Singapore, April 24-28, 2025},
  publisher    = {OpenReview.net},
  year         = {2025},
  url          = {https://openreview.net/forum?id=pZiyCaVuti},
  bibsource    = {dblp computer science bibliography, https://dblp.org}
}

@inproceedings{maharana2024locomo,
  author       = {Adyasha Maharana and
                  Dong{-}Ho Lee and
                  Sergey Tulyakov and
                  Mohit Bansal and
                  Francesco Barbieri and
                  Yuwei Fang},
  editor       = {Lun{-}Wei Ku and
                  Andre Martins and
                  Vivek Srikumar},
  title        = {Evaluating Very Long-Term Conversational Memory of {LLM} Agents},
  booktitle    = {Proceedings of the 62nd Annual Meeting of the Association for Computational
                  Linguistics (Volume 1: Long Papers), {ACL} 2024, Bangkok, Thailand,
                  August 11-16, 2024},
  pages        = {13851--13870},
  publisher    = {Association for Computational Linguistics},
  year         = {2024},
  url          = {https://doi.org/10.18653/v1/2024.acl-long.747},
  doi          = {10.18653/V1/2024.ACL-LONG.747},
  bibsource    = {dblp computer science bibliography, https://dblp.org}
}

@inproceedings{lewis2020rag,
  author       = {Patrick Lewis and
                  Ethan Perez and
                  Aleksandra Piktus and
                  Fabio Petroni and
                  Vladimir Karpukhin and
                  Naman Goyal and
                  Heinrich K{\"{u}}ttler and
                  Mike Lewis and
                  Wen{-}tau Yih and
                  Tim Rockt{\"{a}}schel and
                  Sebastian Riedel and
                  Douwe Kiela},
  editor       = {Hugo Larochelle and
                  Marc'Aurelio Ranzato and
                  Raia Hadsell and
                  Maria{-}Florina Balcan and
                  Hsuan{-}Tien Lin},
  title        = {Retrieval-Augmented Generation for Knowledge-Intensive {NLP} Tasks},
  booktitle    = {Advances in Neural Information Processing Systems 33: Annual Conference
                  on Neural Information Processing Systems 2020, NeurIPS 2020, December
                  6-12, 2020, virtual},
  year         = {2020},
  url          = {https://proceedings.neurips.cc/paper/2020/hash/6b493230205f780e1bc26945df7481e5-Abstract.html},
  bibsource    = {dblp computer science bibliography, https://dblp.org}
}

@article{robertson2009bm25,
  author       = {Stephen E. Robertson and
                  Hugo Zaragoza},
  title        = {The Probabilistic Relevance Framework: {BM25} and Beyond},
  journal      = {Found. Trends Inf. Retr.},
  volume       = {3},
  number       = {4},
  pages        = {333--389},
  year         = {2009},
  url          = {https://doi.org/10.1561/1500000019},
  doi          = {10.1561/1500000019},
  bibsource    = {dblp computer science bibliography, https://dblp.org}
}

@inproceedings{shi2023distracted,
  author       = {Freda Shi and
                  Xinyun Chen and
                  Kanishka Misra and
                  Nathan Scales and
                  David Dohan and
                  Ed H. Chi and
                  Nathanael Sch{\"{a}}rli and
                  Denny Zhou},
  editor       = {Andreas Krause and
                  Emma Brunskill and
                  Kyunghyun Cho and
                  Barbara Engelhardt and
                  Sivan Sabato and
                  Jonathan Scarlett},
  title        = {Large Language Models Can Be Easily Distracted by Irrelevant Context},
  booktitle    = {International Conference on Machine Learning, {ICML} 2023, 23-29 July
                  2023, Honolulu, Hawaii, {USA}},
  series       = {Proceedings of Machine Learning Research},
  volume       = {202},
  pages        = {31210--31227},
  publisher    = {{PMLR}},
  year         = {2023},
  url          = {https://proceedings.mlr.press/v202/shi23a.html},
  bibsource    = {dblp computer science bibliography, https://dblp.org}
}

@article{liu2024lost,
  author       = {Nelson F. Liu and
                  Kevin Lin and
                  John Hewitt and
                  Ashwin Paranjape and
                  Michele Bevilacqua and
                  Fabio Petroni and
                  Percy Liang},
  title        = {Lost in the Middle: How Language Models Use Long Contexts},
  journal      = {Trans. Assoc. Comput. Linguistics},
  volume       = {12},
  pages        = {157--173},
  year         = {2024},
  url          = {https://doi.org/10.1162/tacl\_a\_00638},
  doi          = {10.1162/TACL\_A\_00638},
  bibsource    = {dblp computer science bibliography, https://dblp.org}
}

@inproceedings{chhikara2025mem0,
  author       = {Prateek Chhikara and
                  Dev Khant and
                  Saket Aryan and
                  Taranjeet Singh and
                  Deshraj Yadav},
  editor       = {In{\^{e}}s Lynce and
                  Nello Murano and
                  Mauro Vallati and
                  Serena Villata and
                  Federico Chesani and
                  Michela Milano and
                  Andrea Omicini and
                  Mehdi Dastani},
  title        = {Mem0: Building Production-Ready {AI} Agents with Scalable Long-Term
                  Memory},
  booktitle    = {{ECAI} 2025 - 28th European Conference on Artificial Intelligence,
                  25-30 October 2025, Bologna, Italy - Including 14th Conference on
                  Prestigious Applications of Intelligent Systems {(PAIS} 2025)},
  series       = {Frontiers in Artificial Intelligence and Applications},
  volume       = {413},
  pages        = {2993--3000},
  publisher    = {{IOS} Press},
  year         = {2025},
  url          = {https://doi.org/10.3233/FAIA251160},
  doi          = {10.3233/FAIA251160},
  bibsource    = {dblp computer science bibliography, https://dblp.org}
}

@article{qwen2024qwen25,
  author       = {An Yang and
                  Baosong Yang and
                  Beichen Zhang and
                  Binyuan Hui and
                  Bo Zheng and
                  Bowen Yu and
                  Chengyuan Li and
                  Dayiheng Liu and
                  Fei Huang and
                  Haoran Wei and
                  Huan Lin and
                  Jian Yang and
                  Jianhong Tu and
                  Jianwei Zhang and
                  Jianxin Yang and
                  Jiaxi Yang and
                  Jingren Zhou and
                  Junyang Lin and
                  Kai Dang and
                  Keming Lu and
                  Keqin Bao and
                  Kexin Yang and
                  Le Yu and
                  Mei Li and
                  Mingfeng Xue and
                  Pei Zhang and
                  Qin Zhu and
                  Rui Men and
                  Runji Lin and
                  Tianhao Li and
                  Tingyu Xia and
                  Xingzhang Ren and
                  Xuancheng Ren and
                  Yang Fan and
                  Yang Su and
                  Yichang Zhang and
                  Yu Wan and
                  Yuqiong Liu and
                  Zeyu Cui and
                  Zhenru Zhang and
                  Zihan Qiu},
  title        = {Qwen2.5 Technical Report},
  journal      = {CoRR},
  volume       = {abs/2412.15115},
  year         = {2024},
  url          = {https://doi.org/10.48550/arXiv.2412.15115},
  doi          = {10.48550/ARXIV.2412.15115},
  eprinttype   = {arXiv},
  eprint       = {2412.15115},
  bibsource    = {dblp computer science bibliography, https://dblp.org}
}

@article{qwen2025qwen3,
  author       = {Qwen Team},
  title        = {Qwen3 Technical Report},
  journal      = {CoRR},
  volume       = {abs/2505.09388},
  year         = {2025},
  url          = {https://doi.org/10.48550/arXiv.2505.09388},
  doi          = {10.48550/ARXIV.2505.09388},
  eprinttype   = {arXiv},
  eprint       = {2505.09388},
  bibsource    = {dblp computer science bibliography, https://dblp.org}
}

@inproceedings{kurfali2025conflicting,
    title = "Conflicting Needles in a Haystack: How {LLM}s behave when faced with contradictory information",
    author = {Kurfali, Murathan  and
      {\"O}stling, Robert},
    editor = "Christodoulopoulos, Christos  and
      Chakraborty, Tanmoy  and
      Rose, Carolyn  and
      Peng, Violet",
    booktitle = "Proceedings of the 2025 Conference on Empirical Methods in Natural Language Processing",
    month = nov,
    year = "2025",
    address = "Suzhou, China",
    publisher = "Association for Computational Linguistics",
    url = "https://aclanthology.org/2025.emnlp-main.1742/",
    doi = "10.18653/v1/2025.emnlp-main.1742",
    pages = "34361--34376",
    ISBN = "979-8-89176-332-6",
}

@inproceedings{wallat2026facts,
  author       = {Jonas Wallat and
                  Wolfgang Nejdl and
                  Sandipan Sikdar},
  editor       = {Maria Liakata and
                  Viviane P. Moreira and
                  Jiajun Zhang and
                  David Jurgens},
  title        = {When Facts Change: Temporal Knowledge Conflict Resolution in LLMs},
  booktitle    = {Findings of the Association for Computational Linguistics, {ACL} 2026,
                  San Diego, California, United States, July 2-7, 2026},
  pages        = {2154--2184},
  publisher    = {Association for Computational Linguistics},
  year         = {2026},
  url          = {https://doi.org/10.18653/v1/2026.findings-acl.103},
  doi          = {10.18653/V1/2026.FINDINGS-ACL.103},
  bibsource    = {dblp computer science bibliography, https://dblp.org}
}

@article{chao2026stale,
  author       = {Hanxiang Chao and
                  Yihan Bai and
                  Rui Sheng and
                  Tianle Li and
                  Yushi Sun},
  title        = {{STALE:} Can {LLM} Agents Know When Their Memories Are No Longer Valid?},
  journal      = {CoRR},
  volume       = {abs/2605.06527},
  year         = {2026},
  url          = {https://doi.org/10.48550/arXiv.2605.06527},
  doi          = {10.48550/ARXIV.2605.06527},
  eprinttype   = {arXiv},
  eprint       = {2605.06527},
  bibsource    = {dblp computer science bibliography, https://dblp.org}
}

@article{reddy2026freshness,
  author       = {Vikas Reddy and
                  Sumanth Challaram},
  title        = {Don't Ask the {LLM} to Track Freshness: {A} Deterministic Recipe
                  for Memory Conflict Resolution},
  journal      = {CoRR},
  volume       = {abs/2606.01435},
  year         = {2026},
  url          = {https://doi.org/10.48550/arXiv.2606.01435},
  doi          = {10.48550/ARXIV.2606.01435},
  eprinttype   = {arXiv},
  eprint       = {2606.01435},
  bibsource    = {dblp computer science bibliography, https://dblp.org}
}

\clearpage
\appendix

\section*{Supplementary material}
\addcontentsline{toc}{section}{Supplementary material}
\section{Reproducibility and evaluation protocol}
\label{app:repro}
Code, prompts, raw model outputs, and summary files are available at \url{https://github.com/oleeviyababu/Memory_Conflict}.

\subsection{Dataset and chronological semantics}
All experiments use the \texttt{Conflict\_Resolution} split of MemoryAgentBench. Our main source uses \texttt{factconsolidation\_sh\_6k}, which contains 455 numbered statements and 100 questions. The robustness source is \texttt{factconsolidation\_sh\_32k}, with 2,310 statements and 100 questions. Statement numbers define the chronology: when two statements conflict about the same relation, the higher-numbered statement is treated as the newer version. Because the benchmark facts are synthetic, ordinary world knowledge is not used as the evaluation target. The MemoryAgentBench code repository is released under the MIT License.

\subsection{Models, Ollama, and decoding}
We evaluate \texttt{qwen2.5:7b-instruct} and \texttt{qwen3:14b}; the corresponding public Qwen checkpoints are Apache-2.0 licensed. They were served by Ollama on a remote university GPU system, and our local Python scripts sent requests through the Ollama HTTP chat API over a VPN. The released code omits private hostnames, usernames, and credentials. The full-history 6K smoke test and 100-question full-context baseline use \texttt{num\_ctx}=16384. The question-only baseline, three-question oracle diagnostic, and 6K BM25 retrieval runs use \texttt{num\_ctx}=4096. The final controlled runs and the corrected 32K BM25 replication use temperature 0, seed 42, \texttt{num\_ctx}=2048, and \texttt{num\_predict}=64. Qwen3 additionally sets the top-level API option \texttt{think=false}.

\subsection{Frozen final prompt}
\begin{quote}\small
You are answering a synthetic memory benchmark. Use only the supplied numbered memory statements as the source of truth, even when they contradict ordinary world knowledge. The statements are chronological: higher statement numbers are newer and replace lower conflicting statements about the same fact. Return only the shortest possible answer phrase. Do not explain.
\end{quote}

\subsection{Retrieval and support audits}
BM25 ranks individual statements from the complete history. We add statements in ranked order until adding another statement would exceed the evidence-token budget, then restore the selected statements to chronological order before presenting them to the model. The 6K experiments use budgets of 64, 128, 256, and 512 tokens while the 32K replication uses 64 and 512. After generation, gold-assisted audits uses the benchmark answers to identify the intended latest statement. They diagnose support presence but are not deployable retrieval metrics. At 32K, manual inspection showed that the automatic chain template mislabeled two questions. Direct inspection finds the benchmark-gold statement in all 100 retrieved contexts at both budgets, and the validated 63-question conflict subset has chain-validated latest support in 63/63 contexts at both budgets.

\subsection{Scoring and paired tests}
The official outcome is normalized substring exact match. Predictions are also classified as latest, stale, unknown, or other by comparison with reconstructed update-chain values. We use exact two-sided McNemar tests for paired comparisons of the same benchmark questions under different conditions. Because questions share a single history, these $p$-values describe this benchmark instance rather than a population of independent histories. Appendix~\ref{app:manual} separately documents identified semantic false positives/negatives and mixed-string outputs without silently altering the benchmark metric.

\FloatBarrier
\clearpage
\section{Complete experiment inventory}
\label{app:inventory}
{\footnotesize\setlength{\tabcolsep}{2pt}\begin{longtable}{@{}p{0.06\linewidth}p{0.24\linewidth}p{0.24\linewidth}p{0.1\linewidth}p{0.22\linewidth}p{0.10\linewidth}@{}}
\caption{Complete experiment inventory. Numbering follows the saved result files.}\label{tab:inventory}\\
\toprule ID & Purpose & Setup & Size & Main observation & Status\\\midrule\endfirsthead
\toprule ID & Purpose & Setup & Size & Main observation & Status\\\midrule\endhead
02 & Full-context smoke & Qwen2.5; full 6K history & 5 questions & 2/5 (40\%) & Preliminary\\
03 & First-five evidence audit & Inspect full-history outputs & 5 questions & Gold/latest text present in failed cases & Diagnostic\\
04 & Oracle prompt diagnostic & Question-only / latest-only / all relevant & 3 x 3 & 0/3; 3/3; 3/3 & Diagnostic\\
05 & Full-context baseline & Qwen2.5; full 6K history & 100 calls & 37/100 & Valid baseline\\
06 & Question-only baseline & Qwen2.5; no supplied memory & 100 calls & 19/100 & Valid baseline; prompt differs\\
07 & Full vs. question-only pairing & Paired 05 and 06 & 100 pairs & 19 both; 18 full-only; 0 question-only-only & Diagnostic\\
08-09 & BM25-512 + support audit & Qwen2.5; 6K & 100 calls & 32/100; gold-assisted support 100/100 & Valid + diagnostic audit\\
10 & Gold-assisted isolated context & Qwen2.5; gold-derived one-statement evidence & 100 calls & 97/100 & Diagnostic only; different prompt\\
11-12 & BM25 budget sweep & 64/128/256/512 tokens & 400 calls & 45/40/36/32\% & Valid\\
13 & Budget trajectories and pairs & Paired BM25 conditions & 100 trajectories & 64 vs 512: 14 vs 1; p=0.00098 & Valid diagnostic\\
14 & Update-chain reconstruction & 6K fact templates & 100 questions & 100 resolved; 74 conflict & Gold-assisted diagnostic\\
15 & Conflict stratification & All baseline/retrieval conditions & 100 questions & Failures concentrated in conflict cases & Diagnostic\\
16-17 & Exploratory conflict intervention & Earlier abstaining prompt & 348 calls & 65\% unrelated vs 12.2\% old+latest+unrelated & Exploratory; prompt-sensitive\\
18 & Frozen-prompt pilot & First 10 conflict questions & 40 calls & 10/10 vs 3/10; p=0.0156 & Pilot\\
19 & Final Qwen2.5 intervention & Frozen equal-size conditions & 148 calls & 74/74 vs 33/74; 41 stale flips & Final confirmatory\\
20b & Final Qwen3 intervention & Frozen equal-size; think=false & 148 calls & 74/74 vs 65/74; 9 stale flips & Final robustness\\
21b & 32K BM25 replication & Qwen2.5; budgets 64 and 512 & 200 calls & 69/100 vs 58/100; 18 harmful flips & Final robustness\\
22 & Manual audit & Cross-run qualitative review & 20 cases & 13 sampled flips genuine; 1 metric miss; 2 ambiguities & Final audit\\
\bottomrule
\end{longtable}}
The experimental sequence were designed as diagnostic steps. preliminary baselines suggested that memory helped but it remained limited. Increasing the retrieval budget showed a degradation trend. chain reconstruction localized the trend to conflict questions. The equal-size intervention isolated the effect of the obsolete competing fact. Qwen3 and 32K runs tested the robustness and the manual audit checked semantic and chain-label edge cases.

\FloatBarrier

\section{Preliminary baselines and oracle diagnostics}
\label{app:baselines}
\begin{table}[H]
\centering
\small
\caption{Preliminary and baseline results.}
\label{tab:baseline-all}
\begin{tabular}{lrrrrp{0.28\linewidth}}
\toprule Condition & $N$ & Correct & Acc. (\%) & Prompt tokens & Interpretation\\\midrule
Full-context smoke & 5 & 2 & 40.0 & 6,978--6,981 & Preliminary subset\\
Full 6K context & 100 & 37 & 37.0 & 6,975--6,986 & Valid baseline\\
Question only & 100 & 19 & 19.0 & -- & Different prompt; context-free\\
Gold-assisted isolated & 100 & 97 & 97.0 & 81--103 & Diagnostic; gold-derived evidence and different prompt\\
\bottomrule
\end{tabular}
\end{table}
The full 6K history improves over question only prompting by 18 percentage points. In the paired comparison, 19 questions are answered correctly under both conditions, 18 correct only with the full history, zero correct only without memory, and 63 wrong under both. This comparison provides context but should not be interpreted as a causal effect of memory.

\begin{table}[H]
\centering
\scriptsize
\caption{First-five full-context evidence audit. ``Gold shown" counts literal occurrences displayed by the audit, not relation-specific support.}
\label{tab:first5}
\begin{tabular}{r p{0.31\linewidth} p{0.16\linewidth} p{0.18\linewidth} rr}
\toprule Q & Question & Gold & Model answer & Correct & Gold shown\\\midrule
1 & Which sport is goaltender associated with? & pes\"{a}pallo & ice hockey & no & 3\\
2 & What is Nobuhiro Watsuki famous for? & The Fairly OddParents & Nobuhiro Watsuki is famous for Rurouni Kenshin. & no & 1\\
3 & Which country was rugby union created in? & India & England & no & 5\\
4 & What is Ferdowsi famous for? & Shahnameh & Ferdowsi is famous for Shahnameh. & yes & 2\\
5 & Which university was Joan Didion educated at? & University of California, Berkeley & University of California, Berkeley & yes & 3\\
\bottomrule
\end{tabular}
\end{table}
The first-five inspection provided the earliest sign that the model could answer incorrectly even when the updated value existed in the supplied history. This finding led us to reconstruct relation-specific update chains instead of relying only on whether the answer string appeared somewhere in the history.

\begin{table}[H]
\centering
\small
\caption{Three-question oracle diagnostic.}
\label{tab:oracle-pilot}
\begin{tabular}{lrrr}
\toprule Condition & Correct & Total & Accuracy (\%)\\\midrule
question\_only & 0 & 3 & 0.0\\
latest\_only & 3 & 3 & 100.0\\
all\_relevant\_chronological & 3 & 3 & 100.0\\
\bottomrule
\end{tabular}
\end{table}
In the three-question pilot, the model can answer all the three cases correctly when supplied with the isolated latest statement or the complete relevant chain in chronological order. The later 100-question gold-assisted diagnostic run reaches 97\%, but it used the gold answer to select the evidence and a different prompt from the final frozen experiments. We therefore do not compare its absolute accuracy directly with final controlled conditions.

\FloatBarrier

\section{Complete 6K evidence-budget analysis}
\label{app:budget}
\begin{table}[H]
\centering
\scriptsize
\caption{Complete 6K BM25 budget sweep. Support is the gold-assisted latest-statement diagnostic.}
\label{tab:budget-complete}
\begin{adjustbox}{max width=\linewidth}\begin{tabular}{rrrrrrrr}
\toprule Budget & Acc. & Conflict acc. & No-conflict acc. & Support & Mean stmts. & Mean tokens & Prompt range\\\midrule
64 & 45.0 & 25.7 & 100.0 & 100/100 & 4.17 & 60.23 & 157--174\\
128 & 40.0 & 20.3 & 96.2 & 100/100 & 8.57 & 124.61 & 221--239\\
256 & 36.0 & 16.2 & 92.3 & 100/100 & 16.99 & 251.32 & 348--366\\
512 & 32.0 & 12.2 & 88.5 & 100/100 & 33.60 & 507.44 & 603--624\\
\bottomrule
\end{tabular}\end{adjustbox}
\end{table}

\begin{figure}[t]
\centering
\includegraphics[
    width=0.78\linewidth
]{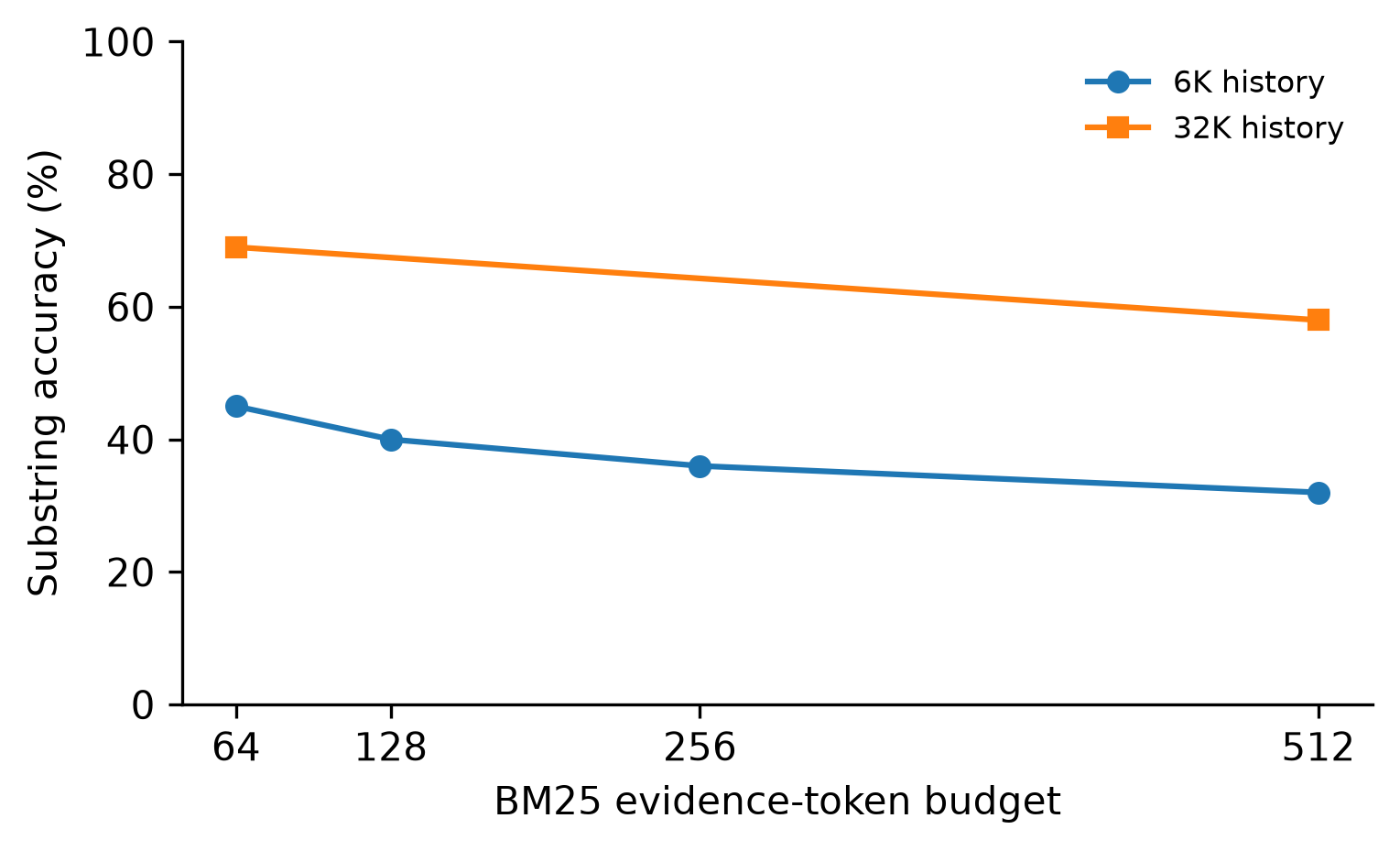}
\caption{Accuracy as a function of BM25 evidence budget. The 6K source provides the full fourbudget sweep, 32K repeats the endpoints.
}
\label{fig:budget}
\end{figure}

\begin{table}[H]
\centering
\small
\caption{All paired 6K BM25 budget comparisons.}
\label{tab:budget-pairs}
\begin{tabular}{rrrrrrrr}
\toprule Lower & Higher & Both correct & Lower-only & Higher-only & Both wrong & Diff. (pt) & Exact $p$\\\midrule
64 & 128 & 36 & 9 & 4 & 51 & 5.0 & 0.266846\\
128 & 256 & 34 & 6 & 2 & 58 & 4.0 & 0.289062\\
256 & 512 & 30 & 6 & 2 & 62 & 4.0 & 0.289062\\
64 & 256 & 33 & 12 & 3 & 52 & 9.0 & 0.035156\\
64 & 512 & 31 & 14 & 1 & 54 & 13.0 & 0.000977\\
128 & 512 & 30 & 10 & 2 & 58 & 8.0 & 0.038574\\
\bottomrule
\end{tabular}
\end{table}
\begin{table}[H]
\centering
\small
\caption{Per-question correctness trajectories across budgets 64, 128, 256, and 512. A one denotes correct.}
\label{tab:trajectories}
\begin{tabular}{lr@{\qquad}lr@{\qquad}lr}
\toprule Trajectory & Count & Trajectory & Count & Trajectory & Count\\\midrule
0000 & 49 & 1111 & 29 & 1000 & 7\\
1110 & 4 & 1100 & 3 & 0100 & 3\\
1001 & 2 & 0010 & 2 & 0111 & 1\\
\bottomrule
\end{tabular}
\end{table}
The monotone aggregate curve does not mean that every individual question worsens. 29 questions are correct at all four budgets, 49 are wrong at all four, and the remaining 22 move between correct and incorrect states. The most informative endpoint comparison has 14 losses and one gain.
{\scriptsize\setlength{\tabcolsep}{2pt}\begin{longtable}{@{}r p{0.25\linewidth} p{0.11\linewidth} p{0.12\linewidth} p{0.12\linewidth} p{0.08\linewidth}@{}}
\caption{All discordant 6K BM25-64 versus BM25-512 questions.}\label{tab:budget-flips}\\
\toprule Q & Question & Gold/latest & Answer 64 & Answer 512 & Direction\\\midrule\endfirsthead
\toprule Q & Question & Gold/latest & Answer 64 & Answer 512 & Direction\\\midrule\endhead
1 & Which sport is goaltender associated with? & pes\"{a}pallo & pes\"{a}pallo & ice hockey & 64-only\\
14 & What is the country of citizenship of C\'{e}dric Klapisch? & Malaysia & Malaysia & France & 64-only\\
18 & Which country was association football created in? & Italy & Italy & England & 64-only\\
24 & Who is the director of American Broadcasting Company? & Ernst Heinkel & Ernst Heinkel & Bob Iger & 64-only\\
25 & Which city did Gennady Rozhdestvensky die in? & United States of America & United States of America & Moscow & 64-only\\
29 & What type of music does Michael Mantler play? & post-punk & post-punk & jazz & 64-only\\
54 & Which religion is Imelda Marcos affiliated with? & atheism & atheism & Catholicism & 64-only\\
55 & What type of music does Dana International play? & Australian hip hop & Australian hip hop & UNKNOWN & 64-only\\
60 & Which sport is Bo Ryan associated with? & association football & basketball & association football & 512-only\\
61 & What type of music does John McVie play? & rap rock & rap rock & rock music & 64-only\\
73 & What position does Blair Walsh play? & placekicker & placekicker & UNKNOWN & 64-only\\
84 & Which religion is Karen Armstrong affiliated with? & Church of Scotland & Church of Scotland & Catholic Church & 64-only\\
94 & What is the country of citizenship of Charles Messier? & Iran & Iran & France & 64-only\\
96 & What position does Mike D'Antoni play? & point guard & point guard & UNKNOWN & 64-only\\
100 & What position does Irving Fryar play? & wide receiver & wide receiver & UNKNOWN & 64-only\\
\bottomrule
\end{longtable}}

\FloatBarrier

\section{Update chains and conflict localization}
\label{app:conflict}
The 6K chain audit successfully reconstructs all 100 questions. Of these, 74 have one older conflicting value, while 26 have no previous conflict, and every chain contains either one or two statements. The update-chain relation templates and the latest statement are identified with gold-assisted post-hoc matching, this analysis is diagnostic rather than deployable.

\begin{table}[H]
\centering
\small
\caption{Complete 6K conflict stratification.}
\label{tab:conflict-complete}
\begin{tabular}{lrrrrr}
\toprule Condition & Conflict correct & Conflict acc. & No-conflict correct & No-conflict acc. & Gap\\\midrule
question\_only & 0/74 & 0.0 & 19/26 & 73.1 & 73.1\\
full\_context & 11/74 & 14.9 & 26/26 & 100.0 & 85.1\\
bm25\_64 & 19/74 & 25.7 & 26/26 & 100.0 & 74.3\\
bm25\_128 & 15/74 & 20.3 & 25/26 & 96.2 & 75.9\\
bm25\_256 & 12/74 & 16.2 & 24/26 & 92.3 & 76.1\\
bm25\_512 & 9/74 & 12.2 & 23/26 & 88.5 & 76.3\\
isolated\_gold & 71/74 & 95.9 & 26/26 & 100.0 & 4.1\\
\bottomrule
\end{tabular}
\end{table}
\begin{table}[H]
\centering
\small
\caption{Paired BM25-64 versus BM25-512 by chain stratum.}
\label{tab:conflict-paired}
\begin{tabular}{lrrrrr}
\toprule Stratum & Both correct & 64-only & 512-only & Both wrong & Net loss\\\midrule
conflict & 8 & 11 & 1 & 54 & 10\\
no\_conflict & 23 & 3 & 0 & 0 & 3\\
\bottomrule
\end{tabular}
\end{table}
The question-only model receives no benchmark memory. It answers none of the 74 synthetic conflict update questions correctly, while it answers 19 of the 26 no-conflict questions correctly, which is consistent with reliance on pretrained knowledge. As the retrieval budget grows, most of the decline occurs in the conflict stratum. Between 64 and 512 tokens, conflict questions show a net loss of 10 correct answers, compared with a net loss of three among no-conflict questions. However, both the latest and old target statements are already present in all 74 conflict contexts at every budget. Therefore, Larger budgets therefore do not newly introduce the obsolete competitor, instead, the results suggest that competing versions make the model more weak when additional evidence is added. The later equal-size intervention tests the effect of the obsolete competitor directly.

\FloatBarrier

\section{Exploratory abstaining-prompt experiment}
\label{app:exploratory}
Before freezing the final prompt, we ran an exploratory four-condition experiment using an earlier prompt that encouraged \texttt{UNKNOWN}. These outputs are useful for understanding prompt sensitivity but we do not treat them as our principal evidence because the high abstention rate obscures stale selection.

\begin{table}[H]
\centering
\small
\caption{Exploratory controlled experiment under the earlier abstaining prompt.}
\label{tab:explore-conditions}
\begin{tabular}{lrrr}
\toprule Condition & Group & Correct / $N$ & Accuracy (\%)\\\midrule
latest\_only & all & 53/100 & 53.0\\
latest\_only & conflict & 28/74 & 37.8\\
latest\_only & no\_conflict & 25/26 & 96.2\\
latest\_plus\_3\_unrelated & all & 65/100 & 65.0\\
latest\_plus\_3\_unrelated & conflict & 39/74 & 52.7\\
latest\_plus\_3\_unrelated & no\_conflict & 26/26 & 100.0\\
old\_plus\_latest & all & 11/74 & 14.9\\
old\_plus\_latest & conflict & 11/74 & 14.9\\
old\_plus\_latest\_plus\_2\_unrelated & all & 9/74 & 12.2\\
old\_plus\_latest\_plus\_2\_unrelated & conflict & 9/74 & 12.2\\
\bottomrule
\end{tabular}
\end{table}
\begin{table}[H]
\centering
\small
\caption{Answer classifications for the 74 conflict questions under the earlier abstaining prompt.}
\label{tab:explore-types}
\begin{tabular}{lrrrr}
\toprule Condition & Latest & Stale & Unknown & Other\\\midrule
latest\_only & 28 & 0 & 46 & 0\\
latest\_plus\_3\_unrelated & 39 & 0 & 34 & 1\\
old\_plus\_latest & 11 & 16 & 45 & 2\\
old\_plus\_latest\_plus\_2\_unrelated & 9 & 27 & 38 & 0\\
\bottomrule
\end{tabular}
\end{table}
The exploratory equal-size comparison, the model received either the latest statement plus three unrelated statements or the old and latest statements plus two unrelated statements yields 31 first-only correct and one second-only correct ($p=1.54\times10^{-8}$). However, the model returned \texttt{UNKNOWN} for 34 questions in the first condition and 38 in the second. The response ``goalie'' for gold ``goaltender'' is categorized as other by substring matching but is a semantic synonym. These findings led us to adopt the final non-abstaining frozen prompt.
\FloatBarrier
\section{Frozen prompt pilot and final cross-model intervention}
\label{app:final}
\begin{table}[H]
\centering
\small
\caption{Frozen-prompt pilot on the first 10 conflict questions.}
\label{tab:frozen-pilot}
\begin{tabular}{lrrr}
\toprule Condition & Correct & Total & Accuracy (\%)\\\midrule
latest\_only & 10 & 10 & 100.0\\
latest\_plus\_3\_unrelated & 10 & 10 & 100.0\\
old\_plus\_latest & 5 & 10 & 50.0\\
old\_plus\_latest\_plus\_2\_unrelated & 3 & 10 & 30.0\\
\bottomrule
\end{tabular}
\end{table}
In the pilot's equal-size comparison has seven latest-plus-unrelated-only successes and no successes in the opposite direction (exact $p=0.015625$). We used this pilot only to check that the non-abstaining prompt produced short, usable answers before the full run.

\begin{table}[H]
\centering
\small
\caption{Final frozen equal-size intervention. Condition A is latest plus three unrelated statements; B is old plus latest plus two unrelated statements.}
\label{tab:final-models}
\begin{tabular}{lrrrrrrrr}
\toprule Model & A acc. & B acc. & Both correct & A-only & B-only & Stale flips & Diff. & Exact $p$\\\midrule
Qwen2.5-7B & 100.0 & 44.6 & 33 & 41 & 0 & 41 & 55.4 & 9.09495e-13\\
Qwen3-14B & 100.0 & 87.8 & 65 & 9 & 0 & 9 & 12.2 & 0.00390625\\
\bottomrule
\end{tabular}
\end{table}
The final Qwen2.5 and Qwen3 experiments each contain 148 completed API calls and no outputs classified as unknown or other.  Within each model, mean prompt-evaluation counts differ by less than one token between A and B, confirming approximately equal prompt size. Under the official classifier, all 50 harmful paired flips are stale selections. Appendix~\ref{app:manual} records two Qwen2.5 mixed-string outputs that are not represented by the four-way label scheme and reports a strict sensitivity analysis.
{\scriptsize\setlength{\tabcolsep}{2pt}\begin{longtable}{@{}p{0.1\linewidth} r p{0.21\linewidth} p{0.12\linewidth} p{0.10\linewidth} p{0.10\linewidth} p{0.10\linewidth}@{}}
\caption{All 50 correct-to-stale flips in the final equal-size interventions.}\label{tab:all-final-flips}\\
\toprule Model & Q & Question & Latest & Obsolete & Answer A & Answer B\\\midrule\endfirsthead
\toprule Model & Q & Question & Latest & Obsolete & Answer A & Answer B\\\midrule\endhead
Qwen2.5-7B & 2 & What is Nobuhiro Watsuki famous for? & The Fairly OddParents & Rurouni Kenshin & The Fairly OddParents & Rurouni Kenshin\\
Qwen2.5-7B & 6 & Who is the author of The Marriage of Figaro? & Thomas Kyd & Pierre Beaumarchais & Thomas Kyd & Pierre Beaumarchais\\
Qwen2.5-7B & 7 & Which sport is Tunisia national football team associated with? & basketball & association football & basketball & association football\\
Qwen2.5-7B & 8 & Which sport is quarterback associated with? & Muay Thai & American football & Muay Thai & American football\\
Qwen2.5-7B & 9 & What is the official language of Japan? & Swedish & Japanese & Swedish & Japanese\\
Qwen2.5-7B & 13 & What position does Robert Parish play? & quarterback & center & quarterback & center\\
Qwen2.5-7B & 14 & What is the country of citizenship of C\'{e}dric Klapisch? & Malaysia & France & Malaysia & France\\
Qwen2.5-7B & 15 & Who is the author of Sidereus Nuncius? & Samuel Beckett & Galileo Galilei & Samuel Beckett & Galileo Galilei\\
Qwen2.5-7B & 16 & Which continent is Germany located in? & Africa & Europe & Africa & Europe\\
Qwen2.5-7B & 17 & Which religion is Kingdom of Ireland affiliated with? & Zoroastrian\-ism & Catholic Church & Zoroastrian\-ism & Catholic Church\\
Qwen2.5-7B & 20 & Who is the chief executive officer of Microsoft? & Steve Jobs & Satya Nadella & Steve Jobs & Satya Nadella\\
Qwen2.5-7B & 30 & Which city did Oscar Wilde die in? & Guangzhou & Paris & Guangzhou & Paris\\
Qwen2.5-7B & 31 & Which country was baseball created in? & Japan & United States of America & Japan & United States of America\\
Qwen2.5-7B & 34 & Which country was rap rock created in? & Russian Empire & United States of America & Russian Empire & United States of America\\
Qwen2.5-7B & 35 & Where was Christianity founded? & Taipei & Jerusalem & Taipei & Jerusalem\\
Qwen2.5-7B & 36 & What is the official language of United States of America? & German & American English & German & American English\\
Qwen2.5-7B & 39 & What is the capital of France? & Harare & Paris & Harare (from statement 171, replaced by a false statement)  
Paris & Paris\\
Qwen2.5-7B & 40 & Who is the developer of SteamOS? & ABB Group & Valve Corporation & ABB Group & Valve Corporation\\
Qwen2.5-7B & 41 & Which city is the headquarter of University of Bucharest located in? & Ankara & Bucharest & Ankara & Bucharest\\
Qwen2.5-7B & 44 & What is the capital of Australia? & Oderzo & Canberra & Oderzo & Canberra\\
Qwen2.5-7B & 45 & Which continent is Malaysia located in? & Antarctica & Asia & Antarctica & Asia\\
Qwen2.5-7B & 49 & Which country was basketball created in? & Soviet Union & United States of America & Soviet Union & United States of America\\
Qwen2.5-7B & 52 & Which sport is Slovakia national football team associated with? & cricket & association football & cricket & association football\\
Qwen2.5-7B & 57 & Which country was Charlie Hebdo created in? & Italy & France & Italy & France\\
Qwen2.5-7B & 58 & Who is Elvis Presley married to? & Charles the Bold & Priscilla Presley & Charles the Bold & Priscilla Presley\\
Qwen2.5-7B & 59 & Who performed Hermione Granger? & Kylie Minogue & Emma Watson & Kylie Minogue & Emma Watson\\
Qwen2.5-7B & 60 & Which sport is Bo Ryan associated with? & association football & basketball & association football & basketball\\
Qwen2.5-7B & 62 & What position does Rog\'{e}rio Ceni play? & flanker & goalkeeper & flanker & goalkeeper\\
Qwen2.5-7B & 64 & What language does Kylie Minogue speak? & German & English & German & English\\
Qwen2.5-7B & 66 & Which sport is placekicker associated with? & rugby & American football & rugby & American football\\
Qwen2.5-7B & 74 & Who is Tony Blair married to? & Alan Jay Lerner & Cherie Blair & Alan Jay Lerner & Cherie Blair\\
Qwen2.5-7B & 78 & Who is the chairperson of Congolese Party of Labour? & Stephen McNeil & Denis Sassou-Nguesso & Stephen McNeil & Denis Sassou-Nguesso\\
Qwen2.5-7B & 81 & What position does Lisa Leslie play? & goaltender & center & goaltender & center\\
Qwen2.5-7B & 82 & What is the name of the current head of state in Soviet Union? & Elizabeth II & Mikhail Gorbachev & Elizabeth II & Mikhail Gorbachev\\
Qwen2.5-7B & 86 & What is the capital of Romania? & Rajanpur & Bucharest & Rajanpur & Bucharest\\
Qwen2.5-7B & 87 & Which continent is India located in? & Africa & Asia & Africa & Asia\\
Qwen2.5-7B & 88 & Which city did Olga of Kiev die in? & Rodez & Kyiv & Rodez & Kyiv\\
Qwen2.5-7B & 89 & Who founded Sasanian Empire? & Walter Chrysler & Ardashir I & Walter Chrysler & Ardashir I\\
Qwen2.5-7B & 91 & Which sport is Surrey Senior Cup associated with? & basketball & association football & basketball & association football\\
Qwen2.5-7B & 92 & What is the country of citizenship of Narendra Modi? & Australia & India & Australia & India\\
Qwen2.5-7B & 99 & Who is Charles Darwin married to? & Amala Paul & Emma Darwin & Amala Paul & Emma Darwin\\
Qwen3-14B & 6 & Who is the author of The Marriage of Figaro? & Thomas Kyd & Pierre Beaumarchais & Thomas Kyd. & Pierre Beaumarchais.\\
Qwen3-14B & 9 & What is the official language of Japan? & Swedish & Japanese & Swedish. & Japanese.\\
Qwen3-14B & 44 & What is the capital of Australia? & Oderzo & Canberra & Oderzo. & Canberra.\\
Qwen3-14B & 57 & Which country was Charlie Hebdo created in? & Italy & France & Italy. & France.\\
Qwen3-14B & 58 & Who is Elvis Presley married to? & Charles the Bold & Priscilla Presley & Charles the Bold. & Priscilla Presley.\\
Qwen3-14B & 59 & Who performed Hermione Granger? & Kylie Minogue & Emma Watson & Kylie Minogue. & Emma Watson.\\
Qwen3-14B & 71 & Which company is AP1000 produced by? & Canadair & Westinghouse Electric & Canadair. & Westinghouse Electric.\\
Qwen3-14B & 89 & Who founded Sasanian Empire? & Walter Chrysler & Ardashir I & Walter Chrysler. & Ardashir I.\\
Qwen3-14B & 92 & What is the country of citizenship of Narendra Modi? & Australia & India & Australia. & India.\\
\bottomrule
\end{longtable}}
\FloatBarrier
\section{Complete 32K replication}
\label{app:sh32k}
The valid 32K experiment contains 200 completed calls. 100 questions at BM25-64 and the same 100 questions with BM25-512. The automated chain audit initially labels 63 conflict and 37 no-conflict questions. Manual review found two questions where the gold-derived chain template attached to an unrelated sentence. We flag these cases as ambiguous rather than silently changing their labels.

\begin{table}[H]
\centering
\small
\caption{Complete 32K BM25 replication. Automated chain support is shown here; direct literal support is 100/100 for both budgets after manual audit.}
\label{tab:32k-complete}
\begin{tabular}{llrrrrrrr}
\toprule Budget & Group & Correct/$N$ & Acc. & Auto support & Latest & Stale & Other & Mean stmts.\\\midrule
64 & all & 69/100 & 69.0 & 98/100 & 69 & 27 & 4 & 3.96\\
64 & conflict & 35/63 & 55.6 & 63/63 & 35 & 27 & 1 & 3.90\\
64 & no\_conflict & 34/37 & 91.9 & 35/37 & 34 & 0 & 3 & 4.05\\
512 & all & 58/100 & 58.0 & 99/100 & 58 & 41 & 1 & 32.96\\
512 & conflict & 22/63 & 34.9 & 63/63 & 22 & 41 & 0 & 32.78\\
512 & no\_conflict & 36/37 & 97.3 & 36/37 & 36 & 0 & 1 & 33.27\\
\bottomrule
\end{tabular}
\end{table}
The official aggregate accuracies are 69/100 (69\%) for BM25-64 and 58/100 (58\%) for BM25-512 . Direct inspection finds the statement supporting the benchmark-gold statement in 100/100 contexts under each budget. After excluding the two ambiguous cases, the clean no-conflict subset contains 35 questions, with accuracies of 34/35 (97.1\%) for BM25-64 and 35/35 (100\%) for BM25-512. Excluding the same two cases only for an overall sensitivity analysis gives $n=98$: 69/98 (70.4\%) versus 57/98 (58.2\%), with 18 BM25-64-only correct, six BM25-512-only correct, 51 both correct, and 23 both wrong (exact $p=0.02266$). These sensitivity values do not replace the official 100-question scores.

{\scriptsize\setlength{\tabcolsep}{2pt}\begin{longtable}{@{}r p{0.28\linewidth} p{0.10\linewidth} p{0.10\linewidth} p{0.10\linewidth} p{0.10\linewidth} p{0.07\linewidth}@{}}
\caption{All 18 detrimental BM25-64 to BM25-512 flips in the 32K replication.}\label{tab:32k-flips}\\
\toprule Q & Question & Latest & Obsolete & Answer 64 & Answer 512 & Type\\\midrule\endfirsthead
\toprule Q & Question & Latest & Obsolete & Answer 64 & Answer 512 & Type\\\midrule\endhead
2 & What position does Nick Rimando play? & flanker & goalkeeper & flanker & goalkeeper & stale\\
12 & What is the country of citizenship of Colin Irwin? & Israel & United Kingdom & Israel & United Kingdom & stale\\
20 & Which religion is Maria Amalia of Naples and Sicily affiliated with? & Church of Greece & Catholic Church & Church of Greece & Catholic Church & stale\\
21 & What position does Tony Parker play? & shooting guard & point guard & shooting guard & point guard & stale\\
42 & Who is the chief executive officer of Apple Inc.? & Vijay Mallya & Tim Cook & Vijay Mallya & Tim Cook & stale\\
48 & Which continent is San Francisco located in? & Africa & North America & Africa & North America & stale\\
54 & Which sport is Imran Khan associated with? & association football & cricket & association football & cricket & stale\\
57 & Who founded Transcendental Meditation movement? & Adalbert of Prague & Maharishi Mahesh Yogi & Adalbert of Prague & Maharishi Mahesh Yogi & stale\\
58 & What is the name of the current head of state in United States of America? & Connachta & Donald Trump & Connachta. & Donald Trump & stale\\
66 & Which religion is Bahawalpur State affiliated with? & Catholic Church & Islam & Catholic Church & Islam & stale\\
67 & Who is the chief executive officer of Advanced Micro Devices? & Sundar Pichai & Lisa Su & Sundar Pichai & Lisa Su & stale\\
69 & Who is the developer of Internet Explorer 5? & Google & Microsoft & Google & Microsoft & stale\\
71 & Who is the author of The Age of Reason? & Allen Ginsberg & Thomas Paine & Allen Ginsberg & Thomas Paine & stale\\
78 & What is the country of citizenship of C\'{e}dric Klapisch? & Malaysia & France & Malaysia & France & stale\\
80 & Which sport is Sports Interactive associated with? & basketball & association football & basketball & association football & stale\\
89 & Who was Finnish created by? & Stirling Silliphanta & Mikael Agricola & Stirling Silliphanta & Mikael Agricola & stale\\
96 & What is the country of citizenship of Henri Gr\'{e}goire? & India & France & India & France & stale\\
99 & What is the name of the current head of the South Korea government? & Mauro Carlesse & Moon Jae-in & Mauro Carlesse & Moon Jae-in & stale\\
\bottomrule
\end{longtable}}
A separate metric audit found one false positive in the 32K results. For the question ``Which sport is flanker associated with?'', the gold answer is ``rugby'', the obsolete value is ``rugby union'', and the model answers ``rugby union'' under both budgets. Because rugby'' is a substring of rugby union'', normalized substring matching counts both responses as correct. Reclassifying this identified case as stale would change the marginal accuracies to 68/100 and 57/100. It would not change the 64-versus-512 paired discordance because the case has the same status under both budgets.

We do not report a full-history 32K baseline because the evidence alone tokenizes to approximately 37.5K tokens. This already exceeds the server model's reported 32,768-token context window before accounting for the system prompt, question, and output allowance.

\FloatBarrier
\section{Manual audit}
\label{app:manual}
We selected twenty cases for the original qualitative audit to cover five Qwen2.5 controlled flips, four Qwen3 controlled flips, four 32K budget flips, two stable correct answers, two exploratory unknown answers, one semantic-scoring case, and two 32K chain ambiguities. The audit is qualitative and does not replace the official substring metric.

{\scriptsize\setlength{\tabcolsep}{2pt}\begin{longtable}{@{}r p{0.15\linewidth} p{0.09\linewidth} r p{0.10\linewidth} p{0.10\linewidth} p{0.09\linewidth} p{0.09\linewidth}@{}}
\caption{Complete 20-case manual audit.}\label{tab:manual20}\\
\toprule ID & Category & Model & Q & Latest & Obsolete & Answer A & Answer B\\\midrule\endfirsthead
\toprule ID & Category & Model & Q & Latest & Obsolete & Answer A & Answer B\\\midrule\endhead
1 & Qwen 2.5 correct-to-stale flip & Qwen2.5 & 2 & The Fairly OddParents & Rurouni Kenshin & The Fairly OddParents & Rurouni Kenshin\\
2 & Qwen 2.5 correct-to-stale flip & Qwen2.5 & 6 & Thomas Kyd & Pierre Beaumarchais & Thomas Kyd & Pierre Beaumarchais\\
3 & Qwen 2.5 correct-to-stale flip & Qwen2.5 & 9 & Swedish & Japanese & Swedish & Japanese\\
4 & Qwen 2.5 correct-to-stale flip & Qwen2.5 & 20 & Steve Jobs & Satya Nadella & Steve Jobs & Satya Nadella\\
5 & Qwen 2.5 correct-to-stale flip & Qwen2.5 & 35 & Taipei & Jerusalem & Taipei & Jerusalem\\
6 & Qwen 3 correct-to-stale flip & Qwen3 & 44 & Oderzo & Canberra & Oderzo. & Canberra.\\
7 & Qwen 3 correct-to-stale flip & Qwen3 & 57 & Italy & France & Italy. & France.\\
8 & Qwen 3 correct-to-stale flip & Qwen3 & 59 & Kylie Minogue & Emma Watson & Kylie Minogue. & Emma Watson.\\
9 & Qwen 3 correct-to-stale flip & Qwen3 & 92 & Australia & India & Australia. & India.\\
10 & 32K BM25 correct-to-stale flip & Qwen2.5 & 2 & flanker & goalkeeper & flanker & goalkeeper\\
11 & 32K BM25 correct-to-stale flip & Qwen2.5 & 12 & Israel & United Kingdom & Israel & United Kingdom\\
12 & 32K BM25 correct-to-stale flip & Qwen2.5 & 42 & Vijay Mallya & Tim Cook & Vijay Mallya & Tim Cook\\
13 & 32K BM25 correct-to-stale flip & Qwen2.5 & 99 & Mauro Carlesse & Moon Jae-in & Mauro Carlesse & Moon Jae-in\\
14 & Stable correct latest answer & Qwen2.5 & 1 & pes\"{a}pallo & ice hockey & pes\"{a}pallo & pes\"{a}pallo\\
15 & Stable correct latest answer & Qwen3 & 3 & India & England & India. & India.\\
16 & Exploratory UNKNOWN / prompt sensitivity & Qwen2.5 & 2 & The Fairly OddParents & Rurouni Kenshin & UNKNOWN & \\
17 & Exploratory UNKNOWN / prompt sensitivity & Qwen2.5 & 9 & Swedish & Japanese & UNKNOWN & \\
18 & Semantic scoring artifact & Qwen2.5 & 81 & goaltender & center & goalie & \\
19 & 32K chain/benchmark ambiguity & Qwen2.5 & 9 & London & Automated chain label was empty & Washington, D.C. & Washington, D.C.\\
20 & 32K chain/benchmark ambiguity & Qwen2.5 & 10 & Rome & Automated chain label was empty & Watertown & Rome\\
\bottomrule
\end{longtable}}
All 13 sampled principal flips involved direct selection of an explicitly stated obsolete value. The final frozen runs contain no answers classified as unknown. The goalie'' response for the gold answer goaltender'' is a metric false negative. The two 32K ambiguities led us to report direct literal support and to exclude from the clean no-conflict subgroup, but they do not alter the official 100-question aggregate scores.

\subsection{Additional metric and mixed-output checks}
A subsequent line-by-line review of the raw outputs identified two further scoring caveats. First, the 32K ``flanker'' answer ``rugby union'' is semantically the reconstructed obsolete value but contains the gold string ``rugby.'' The official substring metric therefore counts it as correct under both budgets. Second, two final Qwen2.5 responses contain both the latest and stale strings. In Condition A, the response to the France-capital question contains Harare \ldots Paris,'' while in Condition B, the response to the \emph{Our Mutual Friend} question is ``Charles Darwin (incorrect, replaced by Charles Dickens).'' Because the gold string appears in each response, the official metric and classifier mark both as latest.

For transparency, a strict sensitivity analysis that treats these two mixed Qwen2.5 responses as non-latest yields 73/74 (98.6\%) in A and 32/74 (43.2\%) in B. The paired comparison remains 41 A-only versus none B-only, so the exact McNemar $p$-value is unchanged. Under this stricter analysis, one of the 41 discordant Condition B responses is mixed rather than a pure stale answer. We retain the official benchmark scores in all primary tables.

\FloatBarrier
\section{Validity and excluded runs}
\label{app:validity}

\subsection{Excluded and diagnostic runs}

We exclude the initial Qwen3 experiment because thinking mode used the
64-token generation allowance before producing a usable final answer.
The valid Qwen3 experiment therefore sets \texttt{think=false} and is
reported separately in the \texttt{20b} result files.

We also exclude the attempted full-history 32K condition. That run used an 8,192-token context setting, while the evidence alone tokenized to approximately 37.5K tokens. The evidence therefore exceeded both the selected context window and the server model's reported maximum context limit of 32,768 tokens, even before accounting for the instructions, question, and output allowance. The reported 32K results therefore use only the corrected BM25-64 and BM25-512 runs from \texttt{21b}.

The gold-assisted isolated and support experiments are used only as
diagnostics because the benchmark answers are used after generation either to select evidence or to identify the intended supporting evidence. The earlier abstaining-prompt experiment is likewise reported only as a prompt-sensitivity analysis and is not combined with the final non-abstaining results.

\subsection{Result verification}

For the canonical runs, we cross-checked the archived raw outputs against their summary files for the reported call counts, accuracies, response classifications, token statistics, paired comparisons, and exact McNemar values. We checked the archived summaries of the excluded runs separately against their recorded configurations and exclusion rationale. We found no inconsistencies in these checks. Private server addresses and credentials are not included
in the archive.
\FloatBarrier

\clearpage
\section{NeurIPS paper checklist}
\section*{NeurIPS Paper Checklist}

\begin{enumerate}

\item {\bf Claims}
    \item[] Question: Do the main claims made in the abstract and introduction accurately reflect the paper's contributions and scope?
    \item[] Answer: \answerYes{} 
    \item[] Justification: The abstract and Introduction state the benchmark setting, controlled intervention, empirical findings, and limits of generalization.

\item {\bf Limitations}
    \item[] Question: Does the paper discuss the limitations of the work performed by the authors?
    \item[] Answer: \answerYes{} 
    \item[] Justification: The Discussion and conclusion section discusses dataset, model, retrieval, metric, statistical, and compute limitations.

\item {\bf Theory assumptions and proofs}
    \item[] Question: For each theoretical result, does the paper provide the full set of assumptions and a complete (and correct) proof?
    \item[] Answer: \answerNA{} 
    \item[] Justification: The paper contains no theoretical theorems or proofs.

    \item {\bf Experimental result reproducibility}
    \item[] Question: Does the paper fully disclose all the information needed to reproduce the main experimental results of the paper to the extent that it affects the main claims and/or conclusions of the paper (regardless of whether the code and data are provided or not)?
    \item[] Answer: \answerYes{} 
    \item[] Justification: Section 2 and Appendix A disclose the dataset sources, prompts, model tags, stage-specific context settings, final inference parameters, retrieval procedure, metrics, and exclusions.

\item {\bf Open access to data and code}
    \item[] Question: Does the paper provide open access to the data and code, with sufficient instructions to faithfully reproduce the main experimental results, as described in supplemental material?
    \item[] Answer: \answerYes{} 
    \item[] Justification: Code, prompts, raw outputs, and summaries are available at \url{https://github.com/oleeviyababu/Memory_Conflict} (Appendix~\ref{app:repro}).

\item {\bf Experimental setting/details}
    \item[] Question: Does the paper specify all the training and test details (e.g., data splits, hyperparameters, how they were chosen, type of optimizer) necessary to understand the results?
    \item[] Answer: \answerYes{} 
    \item[] Justification: Section 2 reports the split, statement and question counts, evidence construction, final model settings, prompt, and evaluation procedure; Appendix A gives stage-specific context settings.

\item {\bf Experiment statistical significance}
    \item[] Question: Does the paper report error bars suitably and correctly defined or other appropriate information about the statistical significance of the experiments?
    \item[] Answer: \answerYes{} 
    \item[] Justification: Section 2 and Appendix A.5 define paired exact McNemar tests, while Sections 3.1--3.3 report discordant cells and exact p-values with an explicit benchmark-diagnostic interpretation.

\item {\bf Experiments compute resources}
    \item[] Question: For each experiment, does the paper provide sufficient information on the computer resources (type of compute workers, memory, time of execution) needed to reproduce the experiments?
    \item[] Answer: \answerNo{} 
    \item[] Justification: The models were served by a university Ollama GPU endpoint whose exact accelerator and quantization metadata were not exposed. Appendix B reports per-stage call counts, while the archived raw JSONL records include client-side timings.
    
\item {\bf Code of ethics}
    \item[] Question: Does the research conducted in the paper conform, in every respect, with the NeurIPS Code of Ethics \url{https://neurips.cc/public/EthicsGuidelines}?
    \item[] Answer: \answerYes{} 
    \item[] Justification: The work uses a public synthetic benchmark, involves no human subjects or private data, and discusses responsible scope and reporting.

\item {\bf Broader impacts}
    \item[] Question: Does the paper discuss both potential positive societal impacts and negative societal impacts of the work performed?
    \item[] Answer: \answerYes{} 
    \item[] Justification: The Discussion and conclusion notes that stale sensitive or operational memories can cause outdated recommendations or disclosures, and that version-aware retirement can mitigate this risk while preserving provenance.
    
\item {\bf Safeguards}
    \item[] Question: Does the paper describe safeguards that have been put in place for responsible release of data or models that have a high risk for misuse (e.g., pre-trained language models, image generators, or scraped datasets)?
    \item[] Answer: \answerNA{} 
    \item[] Justification: The work releases no high-risk model or newly scraped dataset requiring special safeguards.

\item {\bf Licenses for existing assets}
    \item[] Question: Are the creators or original owners of assets (e.g., code, data, models), used in the paper, properly credited and are the license and terms of use explicitly mentioned and properly respected?
    \item[] Answer: \answerYes{} 
    \item[] Justification: Appendix A states that the MemoryAgentBench code repository is MIT licensed and the evaluated Qwen checkpoints are Apache-2.0 licensed; the corresponding benchmark and model papers are cited in the main paper.

\item {\bf New assets}
    \item[] Question: Are new assets introduced in the paper well documented and is the documentation provided alongside the assets?
    \item[] Answer: \answerYes{} 
    \item[] Justification: The new scripts, result records, prompts, and audit protocol are documented in the methods and appendices and are released with a README at \url{https://github.com/oleeviyababu/Memory_Conflict}.

\item {\bf Crowdsourcing and research with human subjects}
    \item[] Question: For crowdsourcing experiments and research with human subjects, does the paper include the full text of instructions given to participants and screenshots, if applicable, as well as details about compensation (if any)? 
    \item[] Answer: \answerNA{} 
    \item[] Justification: The work contains no crowdsourcing or human-subject data collection.

\item {\bf Institutional review board (IRB) approvals or equivalent for research with human subjects}
    \item[] Question: Does the paper describe potential risks incurred by study participants, whether such risks were disclosed to the subjects, and whether Institutional Review Board (IRB) approvals (or an equivalent approval/review based on the requirements of your country or institution) were obtained?
    \item[] Answer: \answerNA{} 
    \item[] Justification: The work contains no human-subject research and therefore requires no IRB review.

\item {\bf Declaration of LLM usage}
    \item[] Question: Does the paper describe the usage of LLMs if it is an important, original, or non-standard component of the core methods in this research? Note that if the LLM is used only for writing, editing, or formatting purposes and does \emph{not} impact the core methodology, scientific rigor, or originality of the research, declaration is not required.
    \item[] Answer: \answerYes{} 
    \item[] Justification: The experimental-design and supplementary reproducibility sections identify the evaluated LLMs and fully specify how they are used as the systems under study.

\end{enumerate}
\end{document}